\documentclass[conference]{IEEEtran}
\usepackage[T1]{fontenc} 
\usepackage[hidelinks]{hyperref}
\usepackage{amsmath}
\usepackage{nicefrac}
\usepackage{tikz}
\usetikzlibrary{tikzmark}
\usepackage{booktabs}
\usepackage{mathtools} 
\usepackage{amsfonts}
\usepackage{longtable}
\usepackage{booktabs}
\usepackage{multirow}
\usepackage{bm}
\usepackage{siunitx}
\usepackage{algorithm}
\usepackage{standalone}
\usepackage{enumerate}
\usepackage{subcaption}
\usepackage{caption}
\usepackage[capitalize]{cleveref}
\crefmultiformat{equation}{Eqs.~(#2#1#3)}%
{ or~(#2#1#3)}{, (#2#1#3)}{ or~(#2#1#3)}
\usepackage[noEnd=true]{algpseudocodex}

\usepackage{vector}

\algrenewcommand\algorithmicfunction{\textbf{Function}}
\usepackage[backend=bibtex,style=ieee,mincitenames=1,maxcitenames=2,maxbibnames=20,url=false,isbn=false,doi=false]{biblatex}

\AtEveryBibitem{\clearfield{month}}

\DeclareFieldFormat{year}{\thefield{year}}
\AtEveryBibitem{%
  \iffieldundef{year}
    {}
    {\clearfield{month}%
     \clearfield{day}%
     \clearfield{endmonth}%
     \clearfield{endday}}%
}

\AtEveryCitekey{%
  \iffieldundef{year}
    {}
    {\clearfield{date}}%
}

\renewbibmacro*{date}{%
  \ifboolexpr{
    test {\iffieldundef{year}}
  }
  {}
  {\printtext{\printfield{year}}}
}
\AtEveryBibitem{\clearfield{series}}

\DeclareBibliographyDriver{unpublished}{%
  \usebibmacro{bibindex}%
  \usebibmacro{begentry}%
  \usebibmacro{author/translator+others}%
  \newunit\newblock
  \usebibmacro{title}%
  \newunit\newblock
  \printfield{note}%
  \newunit\newblock
  \printfield{eprinttype}%
  \newunit
  \printfield{eprint}%
  \newunit\newblock
  \usebibmacro{date}%
  \newunit\newblock
  \usebibmacro{finentry}%
}

\DeclareBibliographyDriver{online}{%
  \usebibmacro{bibindex}%
  \usebibmacro{begentry}%
  \usebibmacro{author/translator+others}%
  \newunit\newblock
  \usebibmacro{title}%
  \newunit\newblock
  \printfield{note}%
  \newunit\newblock
  \printfield{eprinttype}%
  \newunit
  \printfield{eprint}%
  \newunit\newblock
  \usebibmacro{date}%
  \newunit\newblock
  \usebibmacro{finentry}%
}
\AtEveryBibitem{\clearfield{pagetotal}}

\usepackage{tikz}
\usepackage{pgfplots}

\usepgfplotslibrary{statistics}
\usepackage{nicefrac}
\pgfplotsset{compat=1.18}
\usetikzlibrary{positioning}
\usepackage[T1]{fontenc}
\usepackage{xcolor}
\usepackage{siunitx}
\usepackage{bm}

\colorlet{bgcolor}{black!4}
\colorlet{darkgreen}{green!50!black}

\newcommand{\xih}{\bar{\xi}}

\usepackage{algorithm}
\usepackage{bm}
\usepackage{mathrsfs}

\algrenewcommand\algorithmicfunction{\textbf{Function}}

\title{Predictive Uncertainty for Runtime Assurance of a Real-Time Computer Vision-Based Landing System}

\author{\IEEEauthorblockN{Romeo Valentin\IEEEauthorrefmark{1}\IEEEauthorrefmark{4}, Sydney M. Katz\IEEEauthorrefmark{1}, Artur B. Carneiro\IEEEauthorrefmark{1}, \\ Don Walker\IEEEauthorrefmark{2}, and Mykel J. Kochenderfer\IEEEauthorrefmark{1}}
\IEEEauthorblockA{\IEEEauthorrefmark{1}Stanford Intelligent Systems Laboratory, Stanford University, Stanford, CA, 94305\\
\IEEEauthorrefmark{2}A$^3$ by Airbus LLC, Sunnyvale, CA, 94086\\
\IEEEauthorrefmark{4}Email: {romeov@stanford.edu}\\
}}

\begin{document}

\maketitle

\begin{abstract}
Recent advances in data-driven computer vision have enabled robust autonomous navigation capabilities for civil aviation, including automated landing and runway detection. However, ensuring that these systems meet the robustness and safety requirements for aviation applications remains a major challenge.
In this work, we present a practical vision-based pipeline for aircraft pose estimation from runway images that represents a step toward the ability to certify these systems for use in safety-critical aviation applications. Our approach features three key innovations: (i) an efficient, flexible neural architecture based on a spatial Soft Argmax operator for probabilistic keypoint regression, supporting diverse vision backbones with real-time inference; (ii) a principled loss function producing calibrated predictive uncertainties, which are evaluated via sharpness and calibration metrics; and (iii) an adaptation of Residual-based Receiver Autonomous Integrity Monitoring (RAIM), enabling runtime detection and rejection of faulty model outputs. We implement and evaluate our pose estimation pipeline on a dataset of runway images. We show that our model outperforms baseline architectures in terms of accuracy while also producing well-calibrated uncertainty estimates with sub-pixel precision that can be used downstream for fault detection.

\end{abstract}


\section{Introduction}\label{introduction}



Autonomous navigation technologies for civilian aircraft, including autonomous landing and detect-and-avoid applications, have witnessed significant advancements from data-driven models used for computer vision, decision-making, and planning tasks \cite{deanGoldenDecadeDeep2022}. When implementing such systems in safety-critical aviation applications, we must make sure they achieve a sufficient level of safety and robustness. Specifically, these systems must generalize to a wide range of images, run in real-time on board an aircraft, and provide runtime assurance to verify availability of the system. To ensure that these systems meet these requirements, proposals from industry and regulatory bodies have introduced certification frameworks such as variations of the V- or W-shaped life cycle \cite{society2010guidelines,garielFrameworkCertificationAIBased2023,easaConceptsDesignAssurance2020}
and the handling of uncertainty quantification \cite{DINSPEC92005}. However, implementation of real systems fulfilling such requirements remains an ongoing challenge.

\begin{figure}[t]
	\centering
	\includegraphics[width=\linewidth]{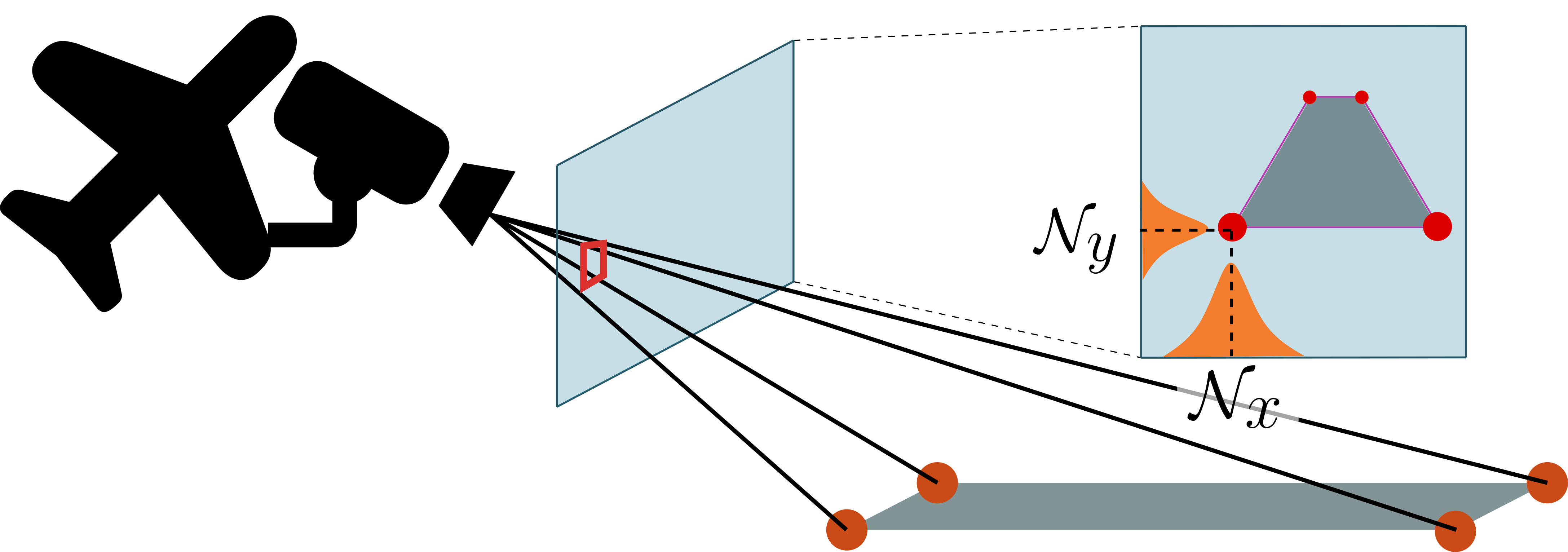}
	\caption{Known world points are projected onto a
		camera, where they are measured as
		under the influence of noise. We wish to determine uncertainties about our measurements and use those to compute our pose and reject outliers.
	}\label{fig:overview}
\end{figure}

In this work, we aim to bridge the gap between regulatory requirements and practical implementation for a specific application: real-time, camera-based pose estimation during a runway approach, paired with a novel outlier detection algorithm for runtime assurance.
We illustrate this application in \cref{fig:overview}.
Specifically, we assume a two-stage pose estimation process where the system (i) precisely locates known runway keypoints (corners, markers, etc) in the image, and (ii) solves a nonlinear optimization problem to determine the pose.
In this paper we focus on the first step and refer to \citeauthor{valentinProbabilisticParameterEstimators2024a} \cite{valentinProbabilisticParameterEstimators2024a} for a discussion of the second step.

We introduce three core technical advancements:

\begin{enumerate}
\item \emph{Efficient Probabilistic Vision Architecture:} We design a lightweight and flexible deep learning architecture for the coordinate regression task.
Using a spatial Soft Argmax operator enables finetuning ``off-the-shelf'' pretrained convolutional neural network (CNN) models to perform sub-pixel precision keypoint regression while adding a minimal parameter footprint.
This approach allows us to leverage recent developments in CNN architectures and pretraining strategies and enables fast inference and real-time use of the system during operation.

  \item \emph{Uncertainty Quantification and Calibration:} Instead of outputting only point estimates, our system outputs dynamic and predictive uncertainty estimates for each detected feature.
		By using the negative log-likelihood loss (a proper scoring rule), and assuming a Gaussian error model, we can train the model to yield well calibrated predictions.
		The uncertainty estimates enable our downstream runtime assurance and can be used for probabilistic pose estimation, see for instance \citeauthor{valentinProbabilisticParameterEstimators2024a} \cite{valentinProbabilisticParameterEstimators2024a}.

  \item \emph{Runtime Integrity Monitoring via RAIM Adaptation:} We adapt Residual-based Receiver Autonomous Integrity Monitoring (RAIM), originally developed for GPS systems, to our vision-based pose estimation pipeline.
  Using the predicted uncertainty estimates, the system can automatically detect and flag predictions that are incompatible with the known runway shape, potentially caused by degraded visibility, corrupted labels, or model failures.
\end{enumerate}

By integrating these components, our pipeline addresses the challenge of runtime assurance detailed by modern certification frameworks.
We emphasize operational feasibility, computational efficiency during training and inference, and enable downstream integration with aviation certification processes.
We validate our approach using the ``Landing Approach Runway Detection'' (LARD) dataset \cite{ducoffe:hal-04056760} to demonstrate the system's efficacy.
Our experiments demonstrate that our system delivers well-calibrated, sharp uncertainty estimates with sub-pixel precision, is easy to build and train with a variety of backbones, supports real-time performance, and can effectively detect and reject faulty model outputs.



\section{Related Work}
Recent breakthroughs in data-driven models have evoked both industry and regulatory interest in developing new autonomous systems.
For example, CNNs have made real-time on-board computer vision algorithms feasible, even in high-speed environments such as autonomous car racing \cite{jaritzEndtoEndRaceDriving2018,rahmanDrivingAutonomyEventBased2024,betzAutonomousVehiclesEdge2022}, drone flying \cite{hanoverAutonomousDroneRacing2024,kaufmannChampionlevelDroneRacing2023,patruno2019vision}, or object tracking \cite{rozantsev2016detecting, james2018learning, opromolla2021visual, ying2022small}.
However, challenges remain in particular when adapting models to large domain shifts or unseen environments \cite{zhouDomainGeneralizationSurvey2023,corsoHolisticAssessmentReliability2023,fangDoesProgressImageNet2023}.
Furthermore, data-driven systems typically are only one element of a larger system and must therefore integrate with a number of other components and their assumptions.
A typical engineering component, especially for state estimation, is the Kalman Filter, which assumes state measurements with Gaussian white noise and known variance.
With these opportunities and requirements in mind, in this work we introduce a network architecture that reliably scales to real-world environments and provides uncertainty estimates for use in downstream tasks.

In addition to designing a robust and generalizable model, we must also build a safety case around it. Certifying traditional aviation systems already requires significant effort, and the introduction of data-driven models into safety-critical applications adds yet another layer of complexity. Recent work has proposed a variety of approaches to address this challenge, including techniques related to formal verification, black-box failure analysis, and runtime assurance.

Formal verification and black-box failure analysis techniques are typically performed offline before deployment. For example, we can use formal verification to prove safety properties of neural network controllers in a variety of domains such as robotics \cite{reachLP,overt,vincent2021reachable} and aviation \cite{Julian2021,manzanas2021verification,Katz2022,bak2022neural}. While formal methods provide guarantees on system safety, they often require strong assumptions and have difficulty scaling to complex, high-dimensional problems. Black-box failure analysis techniques are better equipped to handle this complexity by treating the system as a black box and relying on sample trajectories to efficiently find failures and estimate failure probabilities \cite{corsoSurveyAlgorithmsBlackBox2021}. These approaches have been an important component of the safety case for aviation systems such as the Airborne Collision Avoidance System X (ACAS X) system \cite{Kochenderfer2008cor}, and recent work has proposed techniques to scale these approaches to high-dimensional systems \cite{Moss2024,Delecki2025eras,Delecki2025codit}.

While formal verification and failure analysis provide safety assurances before deployment, it is also important to continue monitoring the safety of the system at runtime \cite{EASA2024AI}. One important component of runtime monitoring is uncertainty quantification. Recent work has proposed a variety of techniques to produce calibrated uncertainty estimates for data-driven systems~\cite{uqreview,conformalprediction,zhaoIndividualCalibrationRandomized2020}. We can use these estimates at runtime to determine the availability of the system. However, one limitation of these predicted uncertainty estimates is that they will be inaccurate in off-nominal scenarios and could cause the aircraft to be confidently misguided by the neural network. To address this limitation, this work adapts a commonly used framework in Global Navigation Satellite System (GNSS) runtime monitoring called RAIM \cite{joergerSolutionSeparationResidualBased2014a,hewitson2006gnss,brown1992baseline,sturza1988navigation} for use in the pose estimation setting. Specifically, we show that we can follow a process similar to the one used to detect faulty satellite pseudoranges in RAIM to detect incorrect neural network predictions.
\section{Preliminaries}\label{sec:preliminaries}
Our vision-based pose estimation pipeline relies on ideas from computer vision, nonlinear optimization, and predictive uncertainty. This section provides the necessary background on these concepts.

\subsection{Pose-from-N-Points (PNP)}\label{subsec:pnp}
To compute an aircraft's pose in relation to the runway, we consider known correspondence points $\{\bm{\xi}_i\}_{i \in 1..K}$ such that we know their real-world three-dimensional coordinates.
If we know the parameters $\mathcal{C}$ of the camera model, we can then estimate our position $\bm{p}$ and rotation parameters $\bm{R}$ by finding the locations $\{\bm{y}_i\}_{i \in 1..K}$ of the correspondence points in the camera image and solving the nonlinear optimization problem
\begin{equation}\arg\min_{(\bm{p}, \bm{R})} \sum_{i}^K{\|\bm{y}_i - \text{project}_{\mathcal{C}}\left( \bm{\xi}, (\bm{p}, \bm{R}) \right)\|^{2}_2}.
\end{equation}
We note that due to the nonlinearity of the projection function this problem can be numerically challenging to solve; in particular for low altitudes, the inverse problem becomes sensitive to small errors in the projection location and its estimation.

One way to counteract the influence of a noisy sample is to weight each summation term by its own uncertainty.
Specifically, if we know the standard deviation $\sigma_i$ of the error for each measurement, choosing $w_i = \sigma_i^{-1}$ for each term provides a principled way to downweight noisy samples without disregarding them completely \cite{DataFittingUncertainty}. In practice, for approach angles of at least \(\qty{1}{\degree}\) above ground, the PNP problem can be solved using the Newton-Raphson, Levenberg-Marquardt, or trust region algorithms \cite{kochenderferAlgorithmsOptimization2019}.

\subsection{Calibration,
	Sharpness, and Proper Scoring Rules}\label{subsec:measuring-calibration-prelims}

Calibration gives us a way to quantify whether the uncertainty in a
series of predictions is faithful to the distribution of the prediction
errors.
Given a set of predicted probability distributions
\(p_{{\xih}_{i}}\) and corresponding measurements \(\xi_{i}\), we
call the predictions \emph{(marginally) calibrated} if
\begin{equation}\text{Pr}\left( \xih_{i} \leq q(p_{\xih_{i}}, \rho) \right) \approx \rho\quad\forall\rho \in (0,1),
	\label{eq:marginal-calibration}
\end{equation}
where \(q(p_{\xih_{i}}, \rho)\) denotes the quantile
function for the random variable \({\xih}_{i}\) evaluated at \(\rho\).
Intuitively, we say that events with predicted probability $\rho$ should come true with frequency $\rho$.

To construct well-calibrated predictions we can minimize a proper scoring rule, which is a loss function that encourages the model to make well-calibrated predictions \cite{gneitingProbabilisticForecastsCalibration2007}.
Notable proper scoring rules include the negative log-likelihood and binary cross-entropy functions.
For further discussion we refer to \citeauthor{gneitingProbabilisticForecasting2014} \cite{gneitingProbabilisticForecasting2014}.

Finally, we note that a series of predictions may be perfectly calibrated yet still be ``bad'' in the sense that each prediction has large uncertainty.
For example, we may predict the probability of rain occurring tomorrow with a single constant prediction through every day of the year, and it will be calibrated if it matches the average number of rainy days.
However, clearly we can hope to do better by also considering the season or weather phenomena. 
We quantify this aspect of a model using a metric called sharpness.
In the case of predicting runway keypoint locations, a similar caveat applies, and we must be sure to make sharp, not just calibrated, predictions.
For the Gaussian distributions discussed in this work, we will simply use the standard deviation of the predicted distributions as a measure of sharpness.



\begin{figure*}
  \centering
  \includegraphics[width=\textwidth]{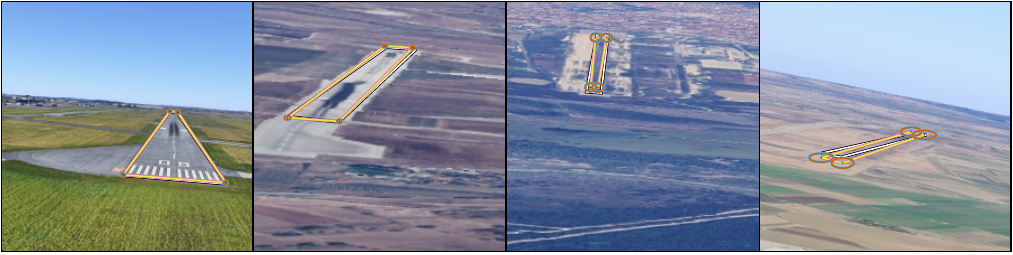}
  \caption{\label{fig:prediction-example}
    Prediction examples with increasing uncertainty. Predicted runway (yellow) overlaid over the true runway (white) and two standard deviations of predicted uncertainty (orange). Each element is contrasted by its opposing color for visual clarity.
  }
\end{figure*}

\section{Methods}
This section presents the three primary components of our vision-based pose estimation pipeline.



\subsection{Coordinate Regression with Soft Argmax}\label{subsec:sam}

Coordinate regression is a fundamental task in computer vision, especially for applications such as keypoint detection, pose estimation, and object localization.
In our scenario, we use coordinate regression to find image coordinates of runway keypoints to estimate our pose with respect to the runway during the approach. Traditional techniques often rely on a CNN backbone. One common approach to coordinate regression involves flattening or pooling the final CNN feature maps and regressing coordinates using fully-connected layers \cite{toshevDeepPoseHumanPose2014}; however, this approach discards the rich spatial structure inherent in convolutional features and may require a large number of extra parameters.
Alternatively, methods may employ complex encoder-decoder architectures with explicit upsampling layers to generate high-resolution heatmaps \cite{ronnebergerUNetConvolutionalNetworks2015,newellStackedHourglassNetworks2016,sunDeepHighResolutionRepresentation2019}. While these preserve spatial detail, the decoder stages significantly increase parameter count and computational cost for both training and inference and can be challenging to train effectively.

To circumvent these trade-offs, we adopt the Soft Argmax (SAM) operator, a lightweight and differentiable mechanism originally proposed by \citeauthor{levineEndtoEndTrainingDeep2016} \cite{levineEndtoEndTrainingDeep2016} to directly extract precise, sub-pixel precision keypoint coordinates from the lower-resolution feature maps produced by the CNN backbone without the need for complex upscaling layers. This approach efficiently leverages the spatial information encoded in the feature maps while allowing end-to-end training and allows us to compute our pose estimate as outlined in \cref{subsec:pnp}.

Specifically, let the CNN backbone take an input image $I \in \mathbb{R}^{H \times W \times C}$ and produce a feature tensor $F \in \mathbb{R}^{H' \times W' \times C'}$, where $H' \ll H, W' \ll W$, and $C' \gg C$. To generate a dedicated spatial map for each of the $K$ target keypoints, we apply a single $1 \times 1$ convolutional layer to $F$. This layer acts as a channel-wise linear transformation, mapping the $C'$ input channels to $K$ output channels, resulting in a tensor $H \in \mathbb{R}^{H' \times W' \times K}$. Each slice $k$ of this tensor, denoted $h_k \in \mathbb{R}^{H' \times W'}$, serves as the activation heatmap for the $k$th keypoint.

The SAM operator then interprets each heatmap $h_k$ probabilistically to compute the expected keypoint position $(\hat{x}_k, \hat{y}_k)$. Conceptually, it finds the location of maximum activation within the heatmap by calculating a softmax across the spatial dimensions, effectively creating a probability distribution $P$ over the grid cells. The final coordinates are the expected values under this distribution, computed as a weighted average of the normalized grid coordinates with 
\begin{align}
    P_{i,j} &= \frac{\exp(h_{k}[i,j])}{\sum_{i'=0}^{H'-1}\sum_{j'=0}^{W'-1} \exp(h_{k}[i',j'])} \label{eq:softmax}\\
    \hat{x}_k &= \sum_{i=0}^{H'-1}\sum_{j=0}^{W'-1} P_{i,j} \cdot \frac{j}{W'-1} \label{eq:sam_x}\\ 
    \hat{y}_k &= \sum_{i=0}^{H'-1}\sum_{j=0}^{W'-1} P_{i,j} \cdot \frac{i}{H'-1} \label{eq:sam_y} 
\end{align}
where $(i, j)$ indexes the spatial dimensions of the heatmap $h_k$. The resulting coordinates $(\hat{x}_k, \hat{y}_k)$ lie in the range $[0, 1]$ and represent continuous locations relative to the heatmap dimensions.
The continuous nature of the output circumvents the issue of the low resolution of $H'$ and $W'$, and the coordinates can be scaled to the original image coordinates when necessary and can achieve sub-pixel precision.

The use of the SAM operator is therefore computationally efficient, requiring only a small number of parameters and computations via the $1 \times 1$ convolution ($(C'+1) \times K$ weights) and the SAM layer (parameter free), avoiding large fully-connected or upsampling layers. 
We note that for a real system, typically a pre-cropping step is required that extracts a crop of the runway from the full resolution image.


\subsection{Predictive Uncertainty via Negative Log Likelihood}\label{subsec:predictive-uncertainties}

While SAM enables efficient and precise coordinate regression, it is essential to characterize the confidence of each prediction for robust downstream usage and reliable system monitoring.
Following \citeauthor{nix1994estimating} \cite{nix1994estimating}, we extend our network to produce not only the spatial coordinates of each keypoint but also an associated predictive uncertainty, which we represent as a standard deviation $\sigma_x$ and $\sigma_y$ for each coordinate.

Therefore, for each keypoint, instead of outputting a point estimate, we predict a Gaussian distribution with a heteroscedastic noise model
\[
    \hat{\bm{y}}_k \sim \mathcal{N}\left(\bm{\mu}_k, \bm{\Sigma}_k\right)
\]
where $\bm{\mu}_k = (\hat{x}_k, \hat{y}_k)$ represents the mean of the predicted distribution for keypoint $k$ and $\bm{\Sigma}_k = \text{diag}\left(\begin{bmatrix}\sigma_{x,k}^2  \sigma_{y,k}^2\end{bmatrix}\right)$ is the diagonal covariance matrix.
Both $\bm{\mu}_k$ and $\bm{\Sigma}_k$ are outputs of the neural network, where $\bm{\mu}_k$ is constructed as discussed in the previous section, and the variance is predicted for each keypoint by pooling and flattening the feature map. We employ a single linear layer that maps from $C'$ to $2K$, and then exponentiate the terms to maintain a positive variance.

As introduced in \cref{subsec:measuring-calibration-prelims}, we enable the network to learn both accurate and well-calibrated uncertainties by optimizing the negative log-likelihood (NLL) loss for the Gaussian model, defined as
\begin{equation}
    \mathcal{L}_{\text{NLL}} = \frac{1}{K} \sum_{k=1}^K \left[\frac{1}{2} \|\bm{L}^{-1}(\bm{y}^{\text{true}}_k - \bm{\mu}_k)\|_2^2 + \log |\bm{\Sigma}_k|\right]
\end{equation}
where $\bm{y}^{\text{true}}_k$ is the ground truth position for keypoint $k \in 1..K$ and $\bm{\Sigma}_k = \bm{L}_k^{\phantom{\top}}\!\bm{L}_k^\top$. This loss function encourages the network to produce large uncertainties for uncertain or hard examples (where the error is large), and it penalizes overconfident predictions through the $\log |\bm{\Sigma}_k|$ term.

\Cref{fig:prediction-example} shows example model predictions on a set of runway images with different levels of predicted uncertainty. By optimizing the probabilistic loss, the model learns to associate each regression output with an explicit, meaningful uncertainty estimate. These predictive uncertainties serve three crucial purposes: they are essential for weighted  pose estimation (\cref{subsec:pnp}), they provide the foundation for runtime integrity and calibration analysis, and they can be used to construct an uncertainty estimate of the pose itself \cite{valentinProbabilisticParameterEstimators2024a}.

The choice of a Gaussian error model is motivated by two main factors.
First, keypoint localization errors in computer vision typically arise from multiple independent sources including camera noise, discretization effects, lighting variations, and model approximation errors.
By the Central Limit Theorem, the superposition of these independent error sources naturally tends toward a Gaussian distribution.
Second, the Gaussian assumption is computationally tractable and enables closed-form solutions for uncertainty propagation in the downstream pose estimation pipeline.
In the experiments presented in this paper we find the Gaussian model to hold well, as is indicated by the good calibration results presented in \cref{fig:calibration} in \cref{sec:experiments}.


\subsection{Runtime Integrity Monitoring through RAIM Adaptation}

\begin{algorithm}[t] 
\caption{Runway Detection Integrity Check}
\label{alg:runway_raim}
\begin{algorithmic}[1] 

\Statex \textbf{Inputs:} Set of $K$ world points $\{ \bm{\xi}_k \}_{k=1..K}$, and
correspondings pixel predictions $\{ (\bm{\mu}_k, \bm{\sigma}_k^2) \}_{k=1..K}$,
camera model $\mathscr{C}$, and rejection threshold $\tau \in [0, 1]$.
\Statex \textbf{Output:} Boolean rejection decision \texttt{ACCEPT} or \texttt{REJECT}.

\Function{checkrejection}{$\{ \bm{\xi}_k \}_k$, $\{ (\bm{\mu}_k, \bm{\sigma}_k^2) \}_k$, $\mathscr{C}$, $\tau$}
\State \text{Estimate pose $(\hat{\bm{p}}, \hat{\bm{R}})$ by solving}
$\min_{\bm{p}, \bm{R}} \sum_{k=1}^{K} \|  \bm{L}_k^{-1} (\text{project}_\mathscr{C}(\bm{\xi}_k, (\bm{p}, \bm{R})) - \bm{\mu}_k) \|_2^2$

where $\bm{\Sigma}_k = \text{diag}(\sigma_{x,k}^2, \sigma_{y,k}^2) = \bm{L}_k^{\phantom{\top}}\!\bm{L}^\top_k$.


\State \text{Reproject points using estimated pose}

$\bm{y}^\text{reproject}_k \gets \text{project}_\mathscr{C}(\bm{\xi}_k, (\hat{\bm{p}}, \hat{\bm{R}}))$ for $k=1..K.$

\State \text{Compute Jacobian $\bm{H} \in \mathbb{R}^{K \times 6}$ of projection function }

$\bm{H}_k \gets \nabla_{(\bm{p}, \bm{R})} \text{project}_\mathscr{C}\left(\bm{\xi}_k, (\bm{p}, \bm{R})\right) \Big|_{(\bm{p}, \bm{R}) = (\hat{\bm{p}}, \hat{\bm{R}})}$

for all rows $k \in 1..K$.

\State \text{Compute residual vector}

$\bm{r} \gets (\bm{I} - \bm{H H^\dagger}) (\bm{y}^{\text{reproject}} - \bm{\mu}).$

where $\bm{H}^\dagger$ is the pseudo-inverse of $\bm{H}$.

\State \text{Compute corrected residual norm}

$\text{stat} \gets \|\bm{L}^{-1} \bm{r}\|_2^2$ 
where $\bm{\Sigma} = \text{diag}(\bm{\Sigma}_1..\bm{\Sigma}_K) = \bm{L}\bm{L^\top}.$

\State \text{Calculate probability density value}

    $p \gets \text{pdf}(\chi^2_{n-m}, \text{stat})$ where $n=2N, m=6.$

\State \text{Make decision:}
\If{$p > \tau$}
    \State \Return \texttt{REJECT}
\Else
    \State \Return \texttt{ACCEPT}
\EndIf

\EndFunction
\end{algorithmic}
\end{algorithm}

\begin{figure}[t]
  \centering
  \includegraphics[width=\linewidth]{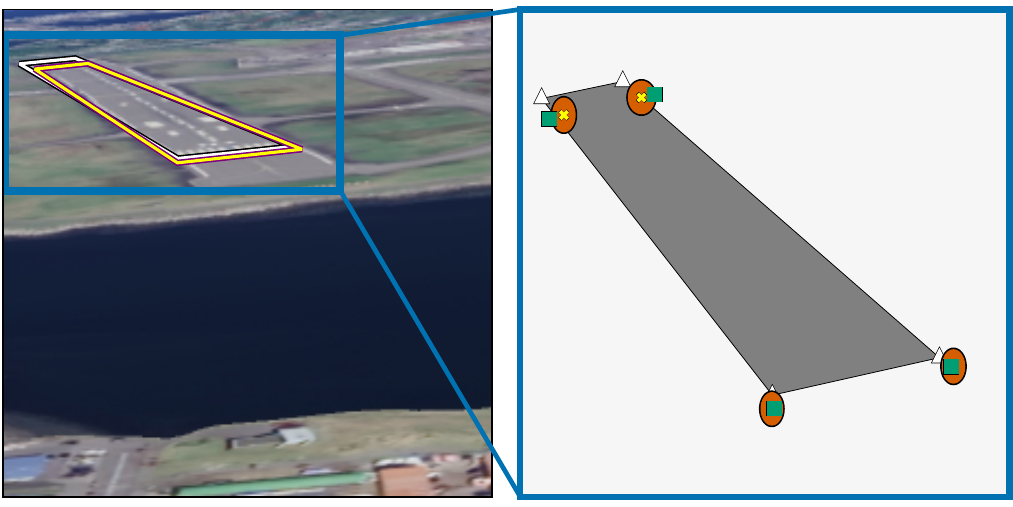}
  \caption{\label{fig:raim-example1}
    Rejection example of a common failure mode: true runway corners (white), corner predictions (yellow), reprojected corners (green) and predicted uncertainty (orange, two standard deviations).}
\end{figure}

To ensure the operational reliability and safety of our pipeline, particularly in safety-critical landing scenarios, it is insufficient to merely produce pose estimates; we must also assess their integrity at runtime. 
It is tempting to rely on the uncertainty outputs directly to assess confidence in a prediction; however, we can only expect our uncertainty estimates to be calibrated on data for which the model generalizes well. Specifically, the model may yield highly confident yet severely incorrect predictions in out-of-distribution scenarios.

For this reason, we adapt the principles of Residual-based RAIM, widely used in GNSS navigation, to validate the integrity of the predicted points and their uncertainties \cite{joergerSolutionSeparationResidualBased2014a}. In GNSS applications, RAIM is used to detect outliers of satellite pseuodorange measurements. We adapt these ideas to perform a similar check for our pixel measurements.



In \Cref{alg:runway_raim}, we outline our proposed integrity check, which uses the entire pose estimation pipeline. After the network predicts keypoint coordinates and their uncertainties, we estimate the aircraft pose by solving a nonlinear least squares problem (\cref{subsec:pnp}). This pose estimate allows us to project the known 3D positions of the runway correspondence points back into the image plane, yielding reprojected keypoint locations $\bm{y}^{\rm proj}_k$.
We then compute the residual error $\bm{r}_k = \bm{\mu}_k - \bm{y}^{\rm proj}_k$ between the model prediction and this geometrically derived expectation for each keypoint.

The predictive uncertainties associated with each output allow us to evaluate the statistical significance of these residuals. 
Under the assumption that the prediction errors follow the predicted Gaussian distributions, the sum of the squared residuals follows a chi-squared ($\chi^2$) distribution after correcting for their predicted variances and accounting for the sensitivity of the projection function to the pose (see step 5 and step 6 in \cref{alg:runway_raim}).
This fact allows us to calculate a test statistic based on the observed residuals and compare it to a threshold derived from the $\chi^2$ distribution given a desired probability of false alarm.
We refer to \citeauthor{joergerSolutionSeparationResidualBased2014a} \cite{joergerSolutionSeparationResidualBased2014a} for mathematical details.

It is important to note that the RAIM adaptation assumes that the prediction errors follow the predicted Gaussian distributions.
In the case of model mismatch the reliability of the integrity monitoring is sensitive to the deviation of predicted uncertainties from the true error characteristics:
If predicted uncertainties are systematically smaller than the true errors, the chi-squared test statistic becomes inflated, leading to excessive alerts that may render the system operationally unusable.
If predicted uncertainties are systematically larger than the true errors, the chi-squared test statistic is less sensitive, allowing more samples to pass but at the cost of higher prediction uncertainties (reduced sharpness).
Well-calibrated uncertainty estimates are therefore crucial for reliable integrity monitoring. 
From a safety perspective, slight underconfidence is generally preferable to overconfidence, as it maintains higher sensitivity to measurement failures at the cost of increased false alarms.


One common situation for which \cref{alg:runway_raim} is particularly effective is when one or more of the keypoint predictions is severely mistaken. 
In \cref{fig:raim-example1} we illustrate an example where the vision model misidentifies the far end of the runway, placing it significantly closer than its actual location (by \SI{184}{\metre}).
No physically plausible aircraft pose will simultaneously satisfy the geometric constraints imposed by both the correct and incorrect keypoints relative to the known runway correspondence points. This inconsistency forces large reprojection errors (residuals) for at least some keypoints when a best-fit pose is computed. Consequently, the overall $\chi^2$ test statistic exceeds the predefined threshold, allowing the system to flag the entire set of measurements as potentially unreliable and triggering an integrity failure alert.

This RAIM-inspired integrity monitor provides a practical and theoretically grounded way to detect problematic model predictions in real time, flagging cases where the network output is inconsistent with physical constraints and learned uncertainties. It represents a crucial component for bridging the gap between data-driven perception and certifiable runtime assurance in aviation applications.


\section{Experiments}\label{sec:experiments}

We assess our pipeline by training and evaluating a series of models on the ``Landing Approach Runway Detection'' (LARD) dataset \cite{ducoffe:hal-04056760}, with \num{10093} train and \num{1261} validation images.
The dataset comprises a variety of runways, lighting conditions, approach angles, and runway distances, thereby representing a reasonable subset of the difficulties of the runway approach problem.
All images are annotated with the pixel coordinates of the four runway corners. We simulate a runway detection system by pre-cropping each image to a $224 \times 224$ crop containing the runway. We train using three standard CNN backbones, each augmented with our SAM head as detailed in \cref{subsec:sam}.
More training details are provided at the end of this section.
We evaluate our trained models in terms of the accuracy, calibration, and ability to detect incorrect outputs. All models run between \num{30} and \SI{60}{\hertz} on a consumer laptop, making them viable for real-time use.








\subsection{Predictive Accuracy}

To evaluate the performance of our architecture, we construct models with three different standard CNN backbones: ResNet18, ResNet50 \cite{heDeepResidualLearning2015} and EfficientNet \cite{tanEfficientNetRethinkingModel2019}. 
For each backbone, we train a baseline model with a standard fully-connected regression head to predict the pixel coordinates of the four runway corners.
We then compare the results to the model introduced in \cref{subsec:sam} constructed and trained with a SAM head on the same task.
We evaluate the predictive accuracy by reporting both the negative log-likelihood (NLL) and the mean absolute pixel error and on a validation set, and we report the results in \Cref{tab:baselinecompare}.

The SAM-based approach outperforms the baseline on both metrics despite its simplicity and minimal parameter count.
Across all three models the SAM-based approach achieves sub-pixel precision with mean absolute pixel errors as low as $0.5$ pixels.
In contrast, the baseline model has significantly larger pixel error and does not work well with the EfficientNet backbone.
We hypothesize that these results are in part due to the SAM layer directly leveraging the spatial structure and rich representations provided by the pretrained backbones and that the minimal parameter count of the model head further benefits generalization to novel inputs since overfitting of the model head becomes less likely.
Finally, we highlight the simplicity of training which and model architecture, since the SAM layer introduces no extra hyperparameters, unlike the baseline model which requires architecting the model head.
For the remainder of our experiments, we use the ResNet18 backbone with the SAM head.

\begin{table}[t]
    \centering
    \caption{Median Performance Results \label{tab:baselinecompare}}
    \begin{tabular}{@{}lcccc@{}}
        \toprule
        \multirow{2}{*}{\textbf{Model}} &
        \multicolumn{2}{c}{\textbf{NLL}} &
        \multicolumn{2}{c}{\textbf{Mean Pixel Error}} \\
        \cmidrule(lr){2-3} \cmidrule(lr){4-5}
        & FC & SAM & FC & SAM \\
        \midrule
        ResNet18       & $-0.068$ & $\bm{-0.079}$ & $1.46$ & $\bm{0.80}$ \\
        ResNet50       & $-0.061$ & $\bm{-0.081}$ & $2.50$ & $\bm{0.65}$ \\
        EfficientNet B5   & $-0.036$ & $\bm{-0.083}$ & $10.59$ & $\bm{0.50}$ \\
        \bottomrule
    \end{tabular}
\end{table}
\subsection{Calibration and Sharpness}
\begin{figure}[t]
    \centering
    \begin{subfigure}[t]{0.48\linewidth}
        \centering
        \includegraphics[scale=0.45]{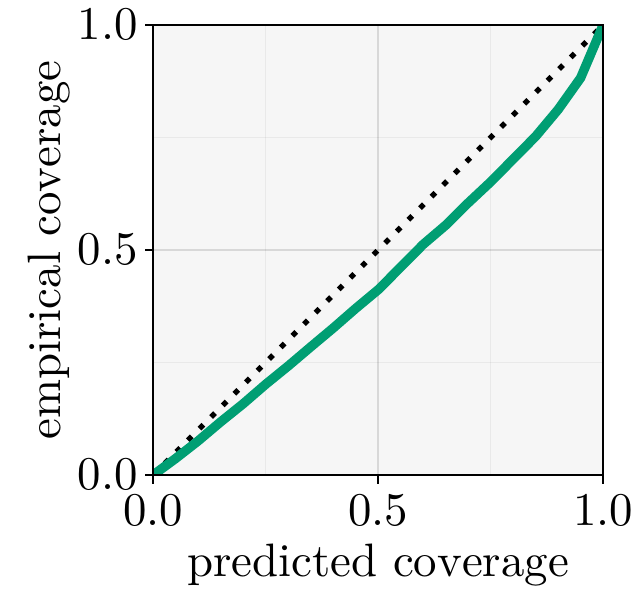}
        \caption{Calibration}
        \label{fig:calibration}
    \end{subfigure}
    ~
    \begin{subfigure}[t]{0.48\linewidth}
        \centering
        \includegraphics[scale=0.45]{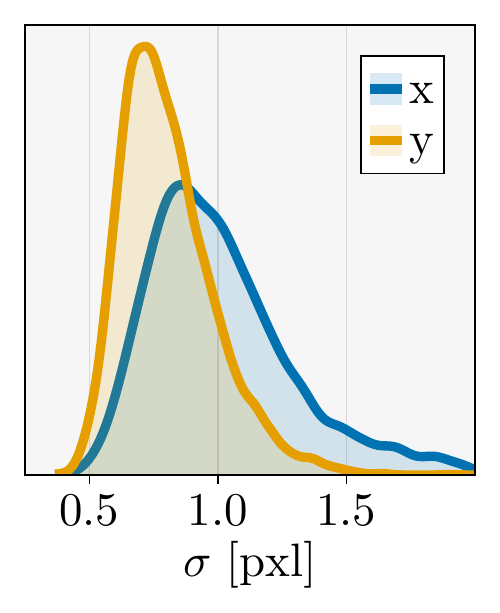}
        \caption{Sharpness}
        \label{fig:sharpness}
    \end{subfigure}
    \caption{Probabilistic metrics for ResNet18 + SAM.} 
    \label{fig:calibration_sharpness} 
\end{figure}

In addition to evaluating the pixel accuracy of our model, we also evaluate its predictive uncertainty estimates and the validate the Gaussian error model assumption.
By plotting the calibration curve, shown in \cref{fig:calibration}, we can assess whether the predicted coverage levels match the observed error frequencies.
In other words, a calibration line close to the identity function validates that our Gaussian model assumption provides a good model for the observed errors.
Indeed, we see that our predictions are well calibrated, justifying the use of this prediction model as an input to the RAIM adaptation.
We note that for application in a safety-critical setting, we may want to apply post-hoc recalibration by increasing each predicted standard deviation by approximately $\SI{20}{\percent}$ to produce a model that is fully calibrated on the validation set.

Measuring precision and calibration is, however, not enough.
As we saw in \cref{subsec:measuring-calibration-prelims}, predictions may be well calibrated but still inaccurate due to an overly large uncertainty.
Therefore, in \cref{fig:sharpness} we plot the sharpness histogram which indicates that the model produces confident (sharp) predictive distributions when appropriate with a typical standard deviation of approximately \num{1} pixel, which matches the result of \cref{tab:baselinecompare}.
We can therefore conclude that the SAM approach introduced in \cref{subsec:sam} not only produces accurate pixel coordinate predictions but also integrates well with the predictive uncertainty framework introduced in \cref{subsec:predictive-uncertainties} and ultimately produces well-calibrated uncertainties for each pixel prediction that fulfill the assumptions required by the RAIM algorithm.



\subsection{Runtime Integrity Monitoring}
\begin{figure}[t]
  \centering
  \includegraphics[width=\linewidth]{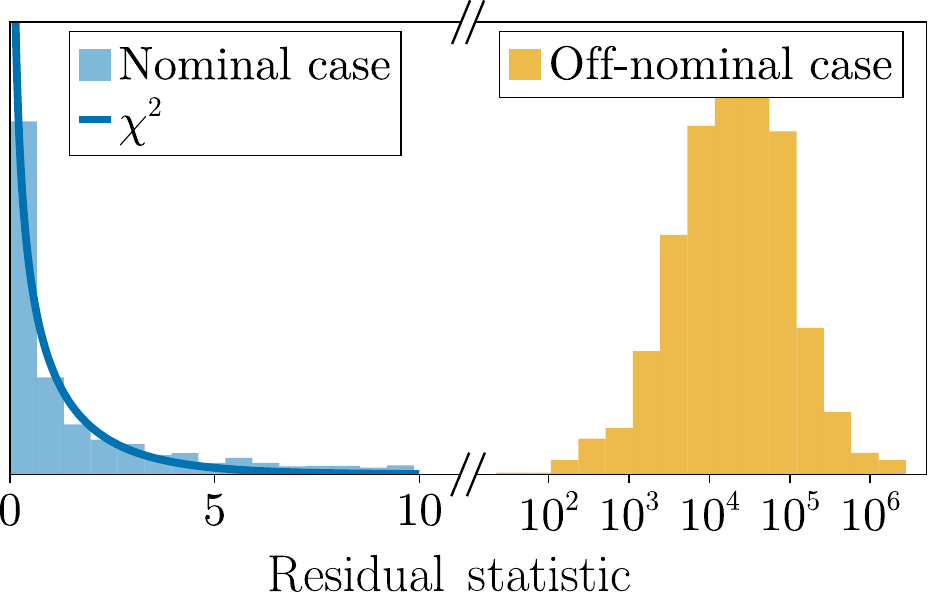}
  \caption{Runtime separation of nominal cases and off-nominal cases with mispredicted far threshold.\label{fig:raim-separation}}
\end{figure}

Finally, to evaluate the efficacy of our runtime integrity monitoring approach, we consider a series of nominal and artificially mispredicted scenarios and examine the residual norm for each case.
For the nominal case, we use our model to predict the keypoint locations for each image in our validation set and use \cref{alg:runway_raim} to compute the residual statistic (step 6).
We plot the results in \cref{fig:raim-separation} against the theoretical $\chi^2$ probability density.
The distribution of the resulting statistics closely matches this theoretical density, which again validates our theoretical assumptions applied to this use case.

However, we are not just interested in whether the residual statistic follows the theoretical distribution in nominal cases. We are also interested in whether off-nominal cases can be reliably detected by the introduced RAIM adaptation such that we can use this method for fault detection and runtime monitoring.
To this end, we compute the residual statistic on artificially mispredicted scenarios as follows. We consider the same model predictions as before but assume that they have been mispredicted in a similar manner to \cref{fig:raim-example1}, where the predicted correspondence points lie \SI{184}{\meter} in front of their true locations.
Specifically, we recompute the residual statistic for all inputs with correspondence points shifted \SI{184}{\meter} behind their true location and plot the results in \cref{fig:raim-separation}.

In this setup, we find that the off-nominal cases are clearly separated from the nominal cases.
In other words, we can reliably detect when a prediction occurs that is incompatible with the geometric constraints of the known correspondence points: no pose can be found such that the reprojections of the known correspondence points lie close to their predicted coordinates relative to the uncertainty.

\subsection{Further Details: Data Processing and Model Training}\label{subsec:experimental-details}
The original images in the dataset have a resolution of approximately 3000x2000 pixels, with the runway often occupying only a small fraction of the image area.
Processing entire image with the model would be computationally very demanding.
To mitigate this issue,
we implement a simple "Bogo crop" strategy: we iteratively sample random 224x224 pixel crops from the image until a crop containing the entire runway (with a specified margin) is found. If the runway dimensions exceed the 224x224 target size in the original image, the image is first repeatedly downsampled by a factor of two until the runway fits within the crop dimensions.

Using these downsampled images, we train each model for 30 epochs over a random subset of 80\% of the LARD training data, and validate on the remaining 20\%.
For each model we match the optimizer proposed in their origina publication: For the ResNet models we use the Adam optimizer with a learning rate of $\num{5e-4}$, and for the EfficientNet model, we use the RMSProp optimizer with a learning rate of $\num{1e-4}$.



\section{Conclusion}
In this work, we have presented a novel computer vision pipeline for robust and fast pose estimation, together with a RAIM-inspired integrity monitoring approach that allows us to do runtime assurance.
We showed how a simple mathematical operator, the Soft Argmax operator, can be used with different CNN backbones to achieve sub-pixel coordinate regression of runway corners or other keypoints in an autonomous landing application.
By using the negative log-likelihood loss function we can construct predictive uncertainties that are well-calibrated, although slightly overconfident, for in-distribution or nominal images.
We then showed that these predictive uncertainties can be used for runtime assurance.

By formulating a mathematically principled description of the residual statistic distribution in nominal cases, we showed that we can clearly detect faulty predictions that are incompatible with the known geometric constraints of the runways.
Notably, the residual statistic can be computed without having access to the ground truth projections and can therefore be computed directly from the network predictions given the correspondence points.
However, we also note that the performance of our RAIM-inspired algorithm degrades as the correlation in the pixel errors of each keypoint increases.
In the limit, if all pixel errors are perfectly correlated, perfect reprojection error can be achieved despite faulty measurements, which makes it impossible for us to detect the fault.

By formulating the runtime assurance problem of a machine learning system as a compatibility problem of its outputs to the known scenario constraints, this approach is applicable to a wide range of machine learning models and estimation problems.
This work therefore represents a concrete step toward bridging the significant gap between regulatory requirements such as EASA's recent guidance for AI systems \cite{easaConceptsDesignAssurance2020} and practical system implementation.
The path to full regulatory compliance, however, requires comprehensive validation of uncertainty calibration and RAIM effectiveness across diverse operational scenarios.
For instance, although the dataset used for validation includes a number of runways, lighting conditions, and approach angles, further studies are necessary to validate the success of our model in real-world images.
Furthermore, the methods introduced in this work can only address a fraction of the overall requirements, and further advancements in uncertainty modeling and integrity monitoring are needed.
We therefore look forward to further work investigating the interplay of the proposed prediction and integrity monitoring approaches with other types of prediction models and real-world problem settings.





\section*{Acknowledgements}
The authors would like to acknowledge Matt Sorgenfrei for his helpful discussions throughout the progression of this work.
This work was supported by A$^3$ by Airbus LLC.
Any opinions, findings, and conclusions expressed in this paper are those of the authors and do not necessarily reflect the views of A$^3$ by Airbus LLC.
\printbibliography

\end{document}
